\documentclass[10pt,twocolumn,letterpaper]{article}

\usepackage{cvpr}              

\usepackage{graphicx}
\usepackage{amsmath}
\usepackage{amssymb}

\usepackage[font=footnotesize,labelfont=bf]{caption}
\usepackage[belowskip=-5pt,aboveskip=3pt]{caption} 

\usepackage{multirow}
\usepackage{tabularx}
\usepackage{dsfont}
\usepackage{url}
\usepackage{color}
\usepackage{amsthm}
\usepackage[export]{adjustbox}
\usepackage{comment}

\usepackage[utf8]{inputenc} 
\usepackage[T1]{fontenc}    
\usepackage{url}            
\usepackage{booktabs}       
\usepackage{amsfonts}       
\usepackage{nicefrac}       
\usepackage{microtype}      
\usepackage{enumitem}
\usepackage{blindtext}
\usepackage{sidecap}

\definecolor{citecolor}{RGB}{65,105,225}

\usepackage{hyperref}
\hypersetup{breaklinks=true,colorlinks,citecolor=citecolor,bookmarks=false}

\usepackage{dsfont}
\definecolor{dg}{rgb}{0,0.694,0.298}
\definecolor{purple}{rgb}{0.4,0.176,0.569}
\definecolor{royalblue}{RGB}{65,105,225}
\usepackage{pifont}
\newcommand{\figref}[1]{Fig.~\ref{#1}}
\newcommand{\reqref}[1]{Eq.~\eqref{#1}}
\newcommand{\secref}[1]{Sec.~\ref{#1}}

\makeatletter
\DeclareRobustCommand\onedot{\futurelet\@let@token\@onedot}
\def\@onedot{\ifx\@let@token.\else.\null\fi\xspace}
\def\eg{\emph{e.g}\onedot} 
\def\ie{\emph{i.e}\onedot}

\def\wrt{w.r.t\onedot} 

\makeatother

\definecolor{americanrose}{rgb}{1.0, 0.01, 0.24}

\usepackage[capitalize]{cleveref}
\crefname{section}{Sec.}{Secs.}
\Crefname{section}{Section}{Sections}
\Crefname{table}{Table}{Tables}
\crefname{table}{Tab.}{Tabs.}

\def\cvprPaperID{3098} 
\def\confName{CVPR}
\def\confYear{2023}

\begin{document}

\title{SuperInpaint: Learning Detail-Enhanced Attentional Implicit Representation for
Super-resolutional Image Inpainting}

\author{Canyu Zhang\textsuperscript{1}\thanks{Xiaoguang Li and Qing Guo are co-first authors and contribute equally.}, 
~Qing Guo\textsuperscript{2\rm$*$}
\thanks{Corresponding author: Qing Guo (\href{mailto:tsingqguo@ieee.org}{tsingqguo@ieee.org})},
~Xiaoguang Li\textsuperscript{1},
~Renjie Wan\textsuperscript{3},\\
~Hongkai Yu\textsuperscript{4},
~Ivor Tsang\textsuperscript{2},
~Song Wang\textsuperscript{1}\\~\\
\textsuperscript{1}University of South Carolina, USA,
\textsuperscript{2}Center for Frontier AI Research (CFAR), A*STAR, Singapore, \\
\textsuperscript{3}Hong Kong Baptist University,
\textsuperscript{4}Cleveland State University, USA\\
}

\twocolumn[{%
\renewcommand\twocolumn[1][]{#1}%
\maketitle

}]


\begin{abstract}

In this work, we introduce a challenging image restoration task, referred to as SuperInpaint, which aims to reconstruct missing regions in low-resolution images and generate completed images with arbitrarily higher resolutions. We have found that this task cannot be effectively addressed by stacking state-of-the-art super-resolution and image inpainting methods as they amplify each other's flaws, leading to noticeable artifacts. To overcome these limitations, we propose the detail-enhanced attentional implicit representation (DEAR) that can achieve SuperInpaint with a single model, resulting in high-quality completed images with arbitrary resolutions.
Specifically, we use a deep convolutional network to extract the latent embedding of an input image and then enhance the high-frequency components of the latent embedding via an adaptive high-pass filter. This leads to detail-enhanced semantic embedding. We further feed the semantic embedding into an unmask-attentional module that suppresses embeddings from ineffective masked pixels. Additionally, we extract a pixel-wise importance map that indicates which pixels should be used for image reconstruction. 
Given the coordinates of a pixel we want to reconstruct, we first collect its neighboring pixels in the input image and extract their detail-enhanced semantic embeddings, unmask-attentional semantic embeddings, importance values, and spatial distances to the desired pixel. 
Then, we feed all the above terms into an implicit representation and generate the color of the specified pixel.
To evaluate our method, we extend three existing datasets for this new task and build 18 meaningful baselines using SOTA inpainting and super-resolution methods. Extensive experimental results demonstrate that our method outperforms all existing methods by a significant margin on four widely used metrics.

\end{abstract}

\section{Introduction}
\label{sec:intro}

%
\begin{figure*}[t]
  \includegraphics[width=\textwidth]{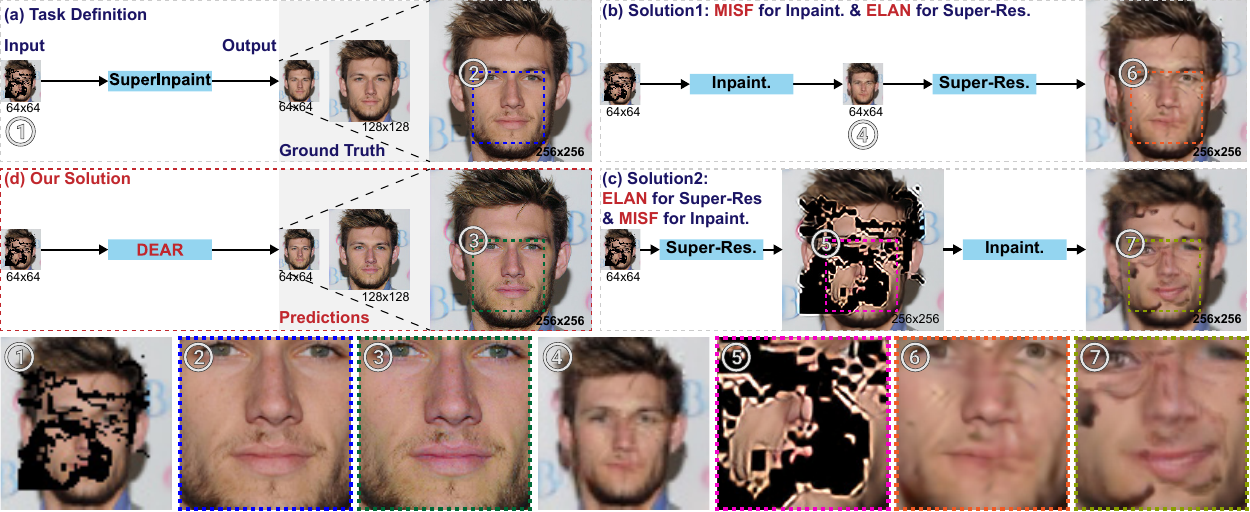} 
  \vspace{-10pt}
  \caption{ Visualization of \emph{SuperInpaint} task (\ie,(a)), two baseline solutions (\ie, (b)~\&~(c)), and our solution (\ie, (d) DEAR). For the two naive solutions, we use the state-of-the-art image inpainting method (\ie, MISF \cite{li2022misf}) and image super-resolution method (\ie, ELAN \cite{zhang2022efficient}). We enlarge seven regions or images at the last row for better visualization.}
  \vspace{-10pt}
  \label{fig:fig1}
\end{figure*}
%

Image inpainting methods \cite{ren2019structureflow, guo2021jpgnet, liu2018image, Dong_2022_CVPR, Liu_2022_CVPR,Lugmayr_2022_CVPR} have been widely used in recent years and achieved promising results, but they still face practical limitations.
Many damaged images are often old and low in resolution. To obtain the best possible view, people often need to complete missing pixels while upscaling these images to higher resolutions. 
The naive choice is to combine the current inpainting and super-resolution (SR) methods directly. However, the performance of such a solution will drop significantly since these two tasks are not directly relevant to each other.
Current SR methods add new pixels based on their nearby features, which can have a negative impact on the entire image when the mask area (\ie, large missing regions) is involved (See \figref{fig:fig1} (c)). Additionally, if the inpainting method generates some flaws, even if they are invisible to the human eye, the SR model can further worsen the results (See \figref{fig:fig1} (b)). 

To overcome these limitations, we focus on a challenging task called \emph{SuperInpaint} that aims to reconstruct missing regions in the input image and generate high-fidelity and completed images with arbitrarily higher resolutions (See \figref{fig:fig1} (a)). \emph{SuperInpaint} has great potential in image inpainting, super-resolution, and old photo restoration areas but cannot be properly addressed by stacking state-of-the-art inpainting and super-resolution methods (See solutions in \figref{fig:fig1} (b) and (c)).  
In this paper, we propose to realize \emph{SuperInpaint} via the implicit image representation \cite{chen2021learning}.
However, \emph{SuperInpaint} has some specific challenges that cannot be properly addressed by the previous work \cite{chen2021learning,lte-jaewon-lee}: First, the arbitrary high-resolution generation requires that the details in the input image should be involved and enhanced in the representation. Second, the effects of the masked pixels on the semantic embedding should be considered. Third, ineffective masked pixels (\ie, unrecoverable masked pixels) should be excluded during the reconstruction process.

To address these challenges, we propose the detail-enhanced attentional implicit representation (DEAR) with three key modules, \ie, the detail-enhanced semantic embedding (DSE) extraction, unmask-attentional semantic embedding (USE) extraction, and importance map extraction.
We first extract the latent embedding of an input image and then use DSE to enhance its high-frequency components.
After that, the embedding is further passed to USE to suppress the information from ineffective masked pixels. 
Moreover, we extract a pixel-wise importance map to indicate which pixels should be used for image reconstruction.
During the testing process, given the coordinates of a pixel we want to reconstruct, we first obtain the pixel's neighboring pixels within the input image and extract the DSE, USE, and importance values of all neighboring pixels as well as their spatial distances to the desired pixel. Then, we feed all the above extractions to the implicit representation (\ie, multi-layer perceptrons (MLPs)) and get the reconstruction result. As shown in \figref{fig:fig1} (d), the proposed method can generate high-fidelity images at different resolutions with rich details.
Overall, the main contributions are as follows:
\begin{itemize}[itemsep=2pt,topsep=0pt,parsep=0pt,leftmargin=*]

    \item We identify a challenging task, \ie, \emph{SuperInpaint} that is to reconstruct missing regions in a low-resolution image and generate high-fidelity and completed images with arbitrarily higher resolutions. 
    
    \item We build 18 baseline methods based on SOTA inpainting and SR methods for this new task and study their effectiveness and limitations.

    \item We propose a novel approach for this task, which is denoted as detail-enhanced attentional implicit representation (DEAR) with novel designed modules.

    \item  We construct new corrupted image datasets in different resolutions based on previous datasets CelebAHQ~\cite{liu2015faceattributes}, Places365~\cite{zhou2017places}, and FFHQ~\cite{karras2019style} by adding masks on the low-resolution counterparts. We conduct extensive experiments on these datasets, which demonstrates the advantages of our method over all existing methods.
\end{itemize}

\section{Related Work}

{\bf Image Inpainting.}
Image inpainting \cite{ zeng2021cr,yu2021frequency, wang2021parallel,yu2021wavefill,suin2021distillation,guo2021image,wang2021image,zhou2021transfill,liao2021image,peng2021generating} aims to restore damaged images by filling in the missing regions according to the non-missing portions. 
Several existing approaches \cite{ren2019structureflow, nazeri2019edgeconnect, liao2020uncertainty} leverage prior information to get realistic results, \eg, \cite{nazeri2019edgeconnect, ren2019structureflow} use edge information and smoothed images to guide the restoration respectively. 
\cite{liao2020uncertainty} performs semantic segmentation and image completion simultaneously to ensure the clear object boundaries. 
In contrast, \cite{liu2018image} employs a novel partial convolution technique that only uses valid pixels to infer the missing pixels. 
%
\cite{guo2021jpgnet} uses an element-wise convolution block to reconstruct missing regions around the mask boundary and a generative network to handle other missing regions. 
\cite{li2022misf} improves upon \cite{guo2021jpgnet} by performing element-wise filtering at both the feature and image levels, where the feature-level filtering focuses on large missing regions and image-level filtering handles local details.
However, traditional image inpainting methods cannot change the image resolution, thus they cannot be deployed on ~\textit{SuperInpaint} task directly.

{\bf Image Super-resolution.}
The goal of SR task~\cite{zhao2022discrete,guo2022lar,yu2022memory,de2022learning,yue2022blind,xu2022dual,kong2022reflash,luo2022learning,yoon2022spheresr} is to increase the image resolution.
Non-learning based methods like Bilinear interpolation add new pixels according to nearby pixels.
Thanks to the development of neural network, 
early regression-based methods such as RCAN~\cite{zhang2018image}, RRDB~\cite{wang2018esrgan}, and EDSR~\cite{lim2017enhanced} can restore the image by learning from the high-level feature and global information.
Then GAN-based methods~\cite{chan2021glean} are proposed based on its generation capacity.
However, those model can only upscale images in certain ratios.
To overcome this limitation, Liif~\cite{chen2021learning} builds a reflection between pixel feature and RGB colour using implicit neural representation, which reconstructs image in arbitrary resolutions.
However, traditional SR models are incapable to deal with the corrupted input in our setting.

{\bf Implicit Neural Representation.}
Implicit neural representations can parameterize a continuous differentiable signal with a neural network, which have been widely used in 2D and 3D tasks~\cite{xiu2022icon, liu2022towards,li2022learning,wu2022neuralhdhair,ye2022gifs,zhao2022self,cao2022jiff}.
Nerf~\cite{mildenhall2021nerf} uses implicit neural field to generate the novel view from differen perspectives. 
IM-NET~\cite{chen2019learning} uses implicit neural function to store the value of point cloud points for 3D shape representation.
LTE~\cite{lte-jaewon-lee} modifies LIIF by adding more high-frequency information in Fourier space, solving the drawback that a standalone MLP is biased toward learning low-frequency knowledge.
And NERV~\cite{NEURIPS2021_b4418237} uses an MLP to store video frames, which is suitable for video compression.
Similar as previous works, we choose a implicit neural representation to build the relationship between pixel information and RGB color. 


\section{\emph{SuperInpaint} and Motivation}
\label{sec:motivation}

\subsection{Problem Formulation}
\label{subsec:formulation}

Given a masked low-resolution image $\mathbf{I}^\text{LR}\in\mathds{R}^{H\times W\times 3}$ where some regions are missing (See Fig.~\ref{fig:fig1}), the task \emph{SuperInpaint} aims to reconstruct the missing pixels in $\mathbf{I}^\text{LR}$ and generate arbitrary higher resolution images denoted as $\mathbf{I}^\text{HR}\in\mathds{R}^{sH\times sW\times 3}$, the variable $s>1.0$ denotes the magnification fraction. The missing regions are indicated by a binary mask tensor $\mathbf{M}\in\mathds{R}^{H\times W}$. 
%

Intuitively, this task could be naively addressed by stacking two existing tasks, \eg, super-resolution (SR) and image inpainting. 
Specifically, we can first use a SR method to handle $\mathbf{I}^\text{LR}$ and get a high-resolution image with missing regions. 
Then, we can further conduct the image inpainting on the high-resolution image to complete the missing pixels.
Similarly, we can also change the execution order and perform image inpainting and super-resolution sequentially. 
To show the challenges of \emph{SuperInpaint} and limitations of the above naive solutions, we select a SOTA image inpainting method (\ie, MISF \cite{li2022misf}) and a SOTA SR method (\ie, ELAN \cite{zhang2022efficient}). Those two methods can consist of two baselines for the \emph{SuperInpaint}: MISF\ding{213}ELAN and ELAN \ding{213} MISF. We analyze their limitations in the following. 

\subsection{Naive Solutions and Limitations}
\label{subsec:limitations}

We take the CelebAHQ dataset \cite{liu2015faceattributes} as an example to construct a dataset for \emph{SuperInpaint}.
We have the training and testing datasets (\ie, $\mathcal{D}_\text{train}$ and $\mathcal{D}_\text{test}$) of CelebAHQ with a high resolution (\ie, $256\times 256$). 
Then, we resize them to low-resolution images (\ie, $64\times 64$) and get datasets $\mathcal{D}_\text{train}^\text{LR}$ and $\mathcal{D}_\text{test}^\text{LR}$.
After that, we mask each image of $\mathcal{D}_\text{train}$, $\mathcal{D}_\text{train}$, $\mathcal{D}_\text{train}^\text{LR}$ and $\mathcal{D}_\text{test}^\text{LR}$ via mask maps from a third-part mask set \cite{liu2018image}, and get the corresponding datasets $\mathcal{C}_\text{train}$, $\mathcal{C}_\text{test}$, $\mathcal{C}_\text{train}^\text{LR}$, and $\mathcal{C}_\text{test}^\text{LR}$.
%
%
We can apply two baselines, \ie, MISF\ding{213}ELAN and ELAN \ding{213} MISF, on the constructed dataset. 

\begin{itemize}[itemsep=2pt,topsep=0pt,parsep=0pt,leftmargin=*]
    \item MISF\ding{213}ELAN: We first train MISF with low-resolution images $\mathcal{D}_\text{train}^\text{LR}$ and the corresponding masked images $\mathcal{C}_\text{train}^\text{LR}$. MISF is fed with a masked low-resolution image and outputs a completed low-resolution image. Then, we train ELAN with low-resolution dataset $\mathcal{D}_\text{train}^\text{LR}$ and the corresponding high-resolution dataset $\mathcal{D}_\text{train}$. We follow default setups in the released codes of MISF and ELAN. During the testing process, we feed an image from $\mathcal{C}_\text{test}^\text{LR}$ to MISF and ELAN sequentially and get a completed higher-resolution image.

    \item ELAN \ding{213} MISF: Similar with MISF\ding{213}ELAN, we first train ELAN with $\mathcal{D}_\text{train}^\text{LR}$ and $\mathcal{D}_\text{train}$. Then, we train MISF with high-resolution images $\mathcal{D}_\text{train}$ and the corresponding masked images $\mathcal{C}_\text{train}$. During the testing process, we feed an image from $\mathcal{C}_\text{test}^\text{LR}$ to ELAN and MISF sequentially and get a completed higher-resolution image.
\end{itemize}

We visualize an example of MISF\ding{213}ELAN and ELAN \ding{213} MISF in \figref{fig:fig1} and see that: \ding{182} MISF\ding{213}ELAN (\ie, \figref{fig:fig1} (b)) can complete the missing pixels with x4 higher resolution. Nevertheless, there are a lot of artifacts around the filled regions of the output (See \figref{fig:fig1} (b)). This is because the super-resolution method is trained to handle normal pixels and cannot properly address the restored pixels, leading to magnified artifacts. \ding{183} ELAN\ding{213}MISF cannot restore natural, enlarged, and completed images. As shown in \figref{fig:fig1} (c), a lot of black regions remain in the final output. The main reason lies in: First, the SOTA super-resolution method (\ie, ELAN \cite{zhang2022efficient}) cannot properly handle the pixels around mask boundaries since it does not see these examples during the training process, leading to distorted and enlarged mask boundaries (See \figref{fig:fig1}). Second, the distorted mask boundaries are further fed to the inpainting method (\ie, MISF\cite{li2022misf}) which also does not see such examples during training, leading to undesired artifacts.

Overall, naively stacking existing inpainting and super-resolution methods cannot address the \emph{SuperInpaint} task since they cannot handle the special patterns (\eg, completed low-resolution pixels or distorted mask boundaries) of intermediate stages. We could retrain the inpainting and super-resolution models to overcome new situations. However, this leads to huge extra training costs and the retraining of one model should be adapted to another one. A novel end-to-end approach is urgently required.


\begin{figure*}[ht]
  \includegraphics[width=\textwidth]{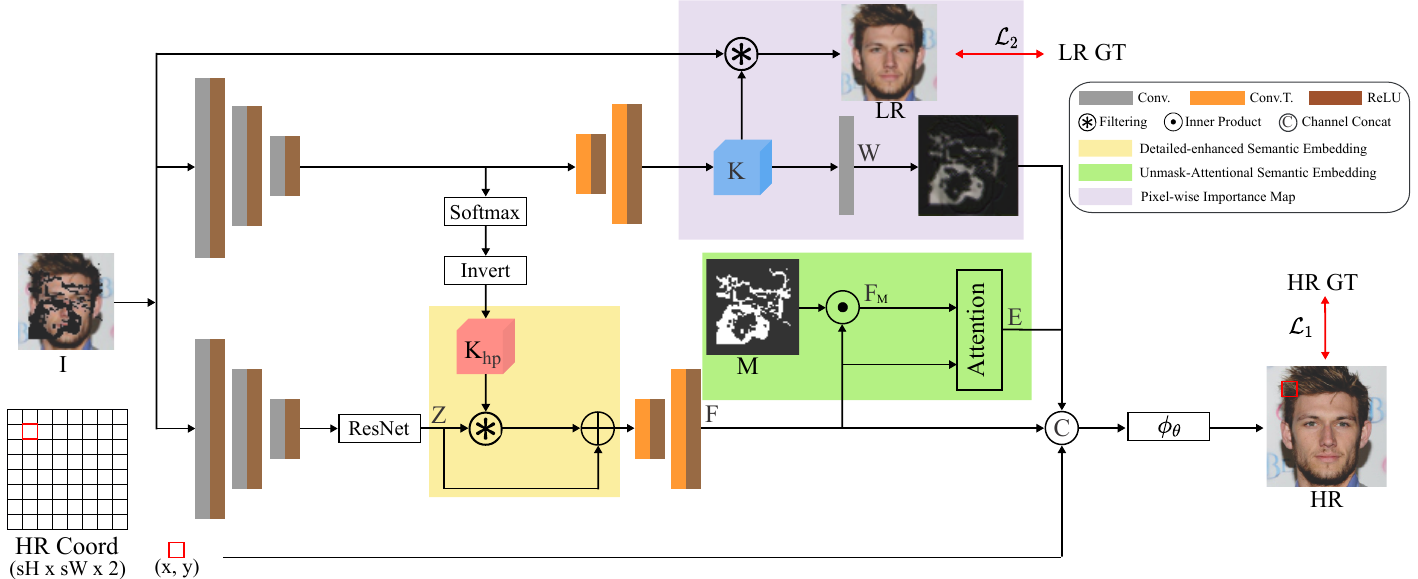} 
  \vspace{-10pt}
  \caption{The overall structure of detail-enhanced attentional implicit representation (DEAR). It is mainly composed of detailed-enhanced semantic embedding extraction, unmask-attentional semantic embedding extraction, and pixel-wise importance map extraction.} 
  \label{main-img}
\end{figure*}

\section{Methodology - DEAR}

\subsection{Overview}

Instead of the two-stage solutions in \secref{sec:motivation}, we propose an end-to-end one-stage method to achieve the \emph{SuperInpaint} task, \ie, detail-enhanced attentional implicit representation (DEAR), which is able to complete missing regions while restoring images in arbitrary resolution.
The intuitive idea is to reconstruct a pixel's color conditional on another pixel based on its detail-enhanced semantic embedding, unmask-attentional embedding, importance, and spatial distance between the two pixels. 
Specifically, given a pixel $\mathbf{p}=(x_p,y_p)$ in the input $\mathbf{I}^\text{LR}\in\mathds{R}^{H\times W\times 3}$, we want to reconstruct the color (\ie, $c_q$) of another pixel at the coordinate $\mathbf{q}=(x_q,y_q)$ that could be continuous coordinated by
%
\begin{align} \label{eq:dear1}
    c_q(\mathbf{p}) = \phi_\theta(\mathbf{F}_\mathbf{p},\mathbf{E}_\mathbf{p},\mathbf{W}_\mathbf{p},\chi(\mathbf{p},\mathbf{q})),
\end{align}
%
where $\mathbf{F}_\mathbf{p}$ denotes the detail-enhanced semantic embedding of the $\mathbf{p}$th pixel, $\mathbf{E}_\mathbf{p}$ is the unmask-attentional embedding of the $\mathbf{p}$th pixel, $\mathbf{W}_\mathbf{p}$ indicates the importance of $\mathbf{p}$th pixel for reconstructing the pixel $\mathbf{q}$, and the function $\chi(\mathbf{p},\mathbf{q})$ calculates the spatial distance between the two pixels. The output $c_q(\mathbf{p})$ represents the color of $\mathbf{q}$th pixel conditional on the $\mathbf{p}$th pixel, and the function $\phi_\theta(\cdot)$ is the desired implicit representation with parameters $\theta$. Note that, once we have a trained \reqref{eq:dear1}, we can calculate the color of arbitrary coordinates. 

Intuitively, we can regard all pixels in $\mathbf{I}^\text{LR}$ as the pixel $\mathbf{p}$ to reconstruct the pixel $\mathbf{q}$. Here, we select spatially neighboring pixels and get the $\mathbf{q}$'s color by
%
\begin{align} \label{eq:dear2}
   c_q = \sum_{\mathbf{p}\in \mathcal{N}_q} w_p c_q(\mathbf{p}),
\end{align}
%
where $\mathcal{N}_q$ denotes the set of neighboring pixels around $\mathbf{q}$, and the weight $w_p$ is determined by spatial distance between $\mathbf{p}$ and $\mathbf{q}$ and normalized to make sure $\sum_{\mathbf{p}\in \mathcal{N}_q} w_p=1$.

The rationales behind the above design are as follows: \ding{182} Previous works \cite{lte-jaewon-lee, abdelmagid2021dynamic} show that deep models like CNNs are usually biased towards learning low-frequency features. However, details (\ie, high-frequency parts) are critical to generate high-fidelity images. Hence, we aim to enhance the detailed information (\eg, high-frequency information) in the semantic embedding and get $\mathbf{F}_\mathbf{p}$ (See \secref{subsec:detailed}). 
\ding{183} The unmasked pixels in the input image are the only reliable information for image restoration. The generative embeddings of masked pixels may affect the implicit representation. As detailed in \secref{subsec:unmask}, we propose to highlight the embeddings of unmasked regions while suppressing the ones of masked pixels. 
\ding{184} If the pixel $\mathbf{p}$ is around the masked regions and can not be recovered accurately, it should not be used to reconstruct the pixel $\mathbf{q}$ in \reqref{eq:dear1}. Hence, we need a weight $\mathbf{W}_\mathbf{p}$ to indicate whether a pixel should be used or not, which is illustrated in \secref{subsec:weight}.
We will validate the effectiveness of all modules in the experimental section.


\subsection{Detailed-enhanced Semantic Embedding}
\label{subsec:detailed}

We propose to enhance the high-frequency parts of the semantic embedding for the implicit representation. Given an encoder $\varphi_\beta(\cdot)$, we extract the latent embedding of the input image $\mathbf{I}$ and get $\mathbf{Z}\in\mathds{R}^{H_z \times W_z\times C_z}$ by $\varphi_\beta(\mathbf{I})$. Then, we aim to extract the high-frequency components of the $\mathbf{Z}$. A straightforward solution is to use high-pass filters to handle each channel of the $\mathbf{Z}$ and add the filtering results back to the input, that is, we have
%
\begin{align}\label{eq:detail1}
   \hat{\mathbf{Z}} = \mathbf{Z}+\mathbf{K}_\text{hp} \circledast \mathbf{Z},
\end{align}
%
where $\mathbf{K}_\text{hp}\in\mathds{R}^{H_z \times W_z\times C_z K_z^2}$ contains the element-wise high-pass filters for each channel. The operation `$\circledast$' denotes element-wise filtering.
The key challenge is how to make the high-pass filters adapt to different inputs and spatial variations. Inspired by \cite{li2022misf, zou2020delving}, we predict  high-pass filters \wrt different inputs
%
\begin{align}\label{eq:detail2}
   \mathbf{K}_\text{hp} = \mathbf{1}-\text{softmax}(\varphi_{\hat{\beta}}(\mathbf{I})),
\end{align}
%
where $\varphi_{\hat{\beta}}(\cdot)$ has a similar architecture with $\varphi_{\beta}(\cdot)$ but with independent parameters and is to predict low-pass filters. Then, we subtract the low-pass filters from the all-one tensor and get high-pass filters.
Note that, we add a softmax layer to the end of $\varphi_{\hat{\beta}}(\cdot)$ to make sure the predicted kernels are low-pass, which makes the predicted kernel weights positive and the summation along channel dimension be one.

After getting the detailed-enhanced latent embedding (\ie, $\hat{\mathbf{Z}}$), we further feed it to a decoder (\ie, $\varphi^{-1}_{\beta^{-1}}(\cdot)$) to get a detailed-enhanced embedding $\mathbf{F}\in\mathds{R}^{H\times W\times C}$ with the same resolution as the input
%
\begin{align}\label{eq:detail3}
   \mathbf{F} = \varphi^{-1}_{\beta^{-1}}(\hat{\mathbf{Z}}),
\end{align}
%
where $\mathbf{F}_\mathbf{p}$ in \reqref{eq:dear1} is the $\mathbf{p}$th element of $\mathbf{F}$.



\subsection{Unmask-attentional Semantic Embedding}
\label{subsec:unmask}

We aim to suppress embeddings of ineffective masked pixels. 
Intuitively, the embeddings of some masked pixels could be recovered by the generative encoder and decoder network (\eg, $\varphi_\beta(\cdot)$ and $\varphi^{-1}_{\beta^{-1}}(\cdot)$) while some masked pixels cannot. We denote the unrecoverable pixels as the ineffective masked pixels whose embeddings should be suppressed.
To this end, we first obtain all unmasked embeddings via the mask $\mathbf{M}$ and $\mathbf{F}$ in \reqref{eq:detail3} and have
%
\begin{align}\label{eq:unmask1}
   \mathbf{F}_\text{M} = (1 - \mathbf{M}) \odot \mathbf{F},
\end{align}
%
where `$\odot$' is the element-wise multiplication. Then, we use the attention operation to calculate weights to measure the ineffective embeddings
%

\begin{align}\label{eq:unmask2}
   \mathbf{E} = \text{softmax}(\frac{\mathbf{F}_\text{M} \otimes  \mathbf{F}^\text{T}}{\sqrt{\text{d}_\text{K}}}) \otimes \mathbf{F},
\end{align}
%
where `$\otimes$' is the matrix production, '$\text{d}_\text{K}$' is the feature channel, and '$\text{T}$' means matrix transpose.
$\mathbf{E}$ reflects the effectiveness of each pixel. 


\subsection{Pixel-wise Importance Map}
\label{subsec:weight}

We also add a new branch to
generate the importance of each pixel, \ie, $\mathbf{W}$.
Specifically, we add two convolution layers (See Fig.~\ref{main-img}) 
to map the latent feature $\varphi_{\hat{\beta}}(\mathbf{I})$ to pixel-wise kernels $\mathbf{K}\in\mathds{R}^{H\times W\times K^2}$ and use these kernels to reconstruct the pixels in the input image. 
For the $\mathbf{p}$th pixel in the input, we can know its $K\times K$ neighboring pixels denoted as the set $\mathcal{N}_p$ and reconstruct the $\mathbf{p}$th pixel by linearly combining the neighboring pixels via the predicted kernels
%
\begin{align}\label{eq:importance1}
   \hat{\mathbf{I}}[\mathbf{p}] = \sum_{\mathbf{q}\in\mathcal{N}_p}\mathbf{K}_\mathbf{p} [\mathbf{p}-\mathbf{q}]\cdot\mathbf{I}[\mathbf{q}],
\end{align}
%
where $\hat{\mathbf{I}}[\mathbf{p}]$ denotes the reconstructed pixel.
Our intuitive idea is that if a pixel could be properly reconstructed via its neighboring pixels, this pixel could be used in the implicit representation and the corresponding importance score $\mathbf{W}_\mathbf{p}$ should be high and towards one.
The predicted kernel can represent the capability of the linear reconstruction and we can generate the importance map by
\begin{align}\label{eq:importance2}
    \mathbf{W} = \text{Conv}(\mathbf{K}).
\end{align}
%

We also visualize the pixel-level importance map in Fig.~\ref{main-img}.
The pixels in the large masked area will be assigned lower importance scores. 
On the contrary, the pixels in unmasked regions and small corrupted regions are given higher importance stores, since small regions can be restored much more easily.
Those results show that proposed importance map can reflect whether certain pixels should be considered during the reconstruction.

\subsection{Implementation Details}
\label{subsec:impl}


{\bf Network architecture.} 
As shown in Fig.~\ref{main-img}, our encoder $\varphi_{\beta}(\cdot)$  contains three convolution layers and eight ResNet blocks.  
For convolution layers, the kernel sizes are 7, 4, and 4. The strides are set to 2. 
The decoder $\varphi^{-1}_{\beta^{-1}}$ is composed of two transposed convolution layers, whose kernel sizes are 4 and strides are 2.
This feature extractor makes sure the generated feature $\text{F}$ has the same resolution as the input.
Our implicit neural function $\phi_\theta$ is parameterized by a four-layer MLP with 256 latent dimensions.

{\bf Loss functions.}
We choose the L1 loss during the training step. 
The predicted pixel color $c_q$ and ground truth pixel color are used to calculate the HR level reconstruction loss $ \mathcal{L}_1$. 
And as illustrated in Sec.~\ref{subsec:weight}, pixel-wise kernel can be applied on corrupted input to get the reconstructed LR image, which is used to calculate the LR level reconstruction loss $  \mathcal{L}_2$.
The final object function is 
\begin{align}
\mathcal{L} = \mathcal{L}_1 + 0.01 * \mathcal{L}_2.
\end{align}

{\bf Hyperparameters.} When training all models, we use Adam as the optimizer ($ \beta_1$= 0.9, $ \beta_2$ = 0.999), the learning rate is 0.0001 and it delays by half every 100 epochs. 
All models are trained for 200 epochs on two NVIDIA Tesla V100 GPUs and the batch size is set to 16.

\section{Experimental Results}

\subsection{Setups}

\begin{table*}[t]
	\centering
	 
	\renewcommand\arraystretch{1.0}
	\footnotesize
	\setlength{\tabcolsep}{1.0mm}{
		\begin{tabular}{l|c|ccccccccccccccccc|c}
			\toprule
   & \multirow{2}{*}{Metric} & \multicolumn{4}{c}{MISF\ding{213}} & BI & EDSR& LTE & ELAN & \multicolumn{4}{c}{LAMA\ding{213}}  & BI & EDSR &  LTE &  ELAN  & \multirow{2}{*}{LTE}& \multirow{2}{*}{DEAR} \\ 
    
      &  & BI & EDSR & LTE & ELAN & \multicolumn{4}{c}{\ding{213}MISF} & BI & EDSR & LTE & ELAN  & \multicolumn{4}{c}{\ding{213}LAMA} &  &  \\ 
			\midrule
			\multirow{4}{*}{\rotatebox{90}{CelebAHQ}} &PSNR  $\uparrow$
& 25.55 & 26.83& 26.09 & 26.89 & 26.28 & 22.83 & 22.65 & 21.06 & 25.70 & 26.36 & 25.48 & 26.41  & 25.03& 22.59 & 22.17& 22.63& 28.26&\textbf{29.84} \\
         
& SSIM$\uparrow$   & 0.871 & 0.872& 0.829 & 0.873 & 0.878 & 0.840 & 0.781 & 0.766 & 0.831 & 0.857 & 0.840 & 0.856 & 0.818& 0.829 &0.758 &0.814 &0.855 &\textbf{0.968} \\

& L1  $\downarrow$ &  0.037 & 0.053 & 0.041 & 0.031 & 0.057 & 0.081 & 0.070 & 0.086 & 
               0.034 & 0.030 & 0.042 & 0.031 & 0.090& 0.184& 0.423 & 0.106 &0.025&\textbf{0.022} \\

& LPIPS $\downarrow$   & 0.331 & 0.143 & 0.070 & 0.141 & 0.413 & 0.704 & 0.070 & 0.273 & 
               0.214  & 0.175 & 0.204 & 0.175 & 0.325& 0.414 & 0.518 & 0.328 &0.236&\textbf{0.053} \\     
                 \midrule

               \multirow{4}{*}{\rotatebox{90}{Places365}} & PSNR $\uparrow$ & 20.81 & 21.49 & 21.29 & 22.08 & 19.47 & 17.97 & 19.04 &18.47 & 
               21.56 & 21.55 & 21.93 & 22.43 & 18.82 & 17.88& 18.27&18.36&23.87 & \textbf{25.61}\\
    
             &  SSIM$\uparrow$ & 0.626 & 0.645& 0.704 & 0.635 & 0.546 & 0.630 & 0.587 & 0.600 & 
               0.682 & 0.702 & 0.622 & 0.758 &0.569 & 0.527& 0.587 &0.581& 0.722&\textbf{0.824} \\
  
                &L1  $\downarrow$  & 0.068 &0.090 & 0.065 & 0.094 & 0.074 & 0.070 & 0.064 & 0.313 & 
               0.059 & 0.056 & 0.054 &0.049 & 0.093& 0.178& 0.273 &0.091 &0.041 & \textbf{0.040}\\
    
             &   LPIPS $\downarrow$    & 0.307 & 0.205 & 0.250 & 0.311 & 0.609 & 0.487 & 0.370 & 0.545 & 
               0.341 & 0.347 & 0.475 & 0.287 &0.430 & 0.522& 0.514 &0.380& 0.354& \textbf{0.170} \\

               \midrule
               \multirow{4}{*}{\rotatebox{90}{FFHQ}} & 		PSNR $\uparrow$ & 24.48 & 25.13 & 24.95 & 25.26 &22.10& 19.60 & 19.29 & 20.28 & 
               24.19 & 24.86 & 23.16 & 24.84 & 22.30& 20.32 & 20.35 &20.84&27.93& \textbf{28.97}\\
             &  SSIM$\uparrow$ & 0.842 & 0.861 & 0.777 & 0.866 &0.821 & 0.749 & 0.712 & 0.786  & 
               0.821 & 0.847 & 0.828 & 0.844&0.824& 0.709 &0.751 &0.752 &0.851  & \textbf{0.879}\\
                &L1  $\downarrow$   & 0.037 & 0.034 & 0.047 & 0.033 &0.108& 0.252 & 0.250  & 0.320  & 
               0.037 & 0.033 & 0.044 & 0.034 & 0.088 & 0.195 & 0.416 &0.100 &0.026 & \textbf{0.023}\\

             &   LPIPS $\downarrow$   & 0.200 & 0.168 & 0.359 & 0.164 &0.462& 0.573  & 0.569 & 0.448 & 
               0.232 & 0.191 & 0.323 & 0.196 & 0.334 & 0.429 & 0.542 &0.303&0.246 & \textbf{0.160}\\
               
			\bottomrule
	\end{tabular}}

        \vspace{5pt}

 	\caption{Results from DEAR and proposed baselines on CelebAHQ~\cite{karras2017progressive}, Places365~\cite{zhou2017places}, and FFHQ datasets~\cite{karras2019style}. 
  The baselines are generated using previous inpainting models (MISF~\cite{li2022misf} and LAMA~\cite{suvorov2021resolution}), and SR models (Bilinear Interpolation(BI), EDSR~\cite{lim2017enhanced}, LTE~\cite{lte-jaewon-lee}, and ELAN~\cite{zhang2022efficient}). }
     \label{main-result}
\end{table*}

\begin{table*}[t]
\vspace{3pt}
	\centering
	
	\renewcommand\arraystretch{1.0}
	\footnotesize
	\setlength{\tabcolsep}{1.0mm}{
		\begin{tabular}{l|cccccccc}
			\toprule
                 \multirow{2}{*}{Method} &  \multicolumn{8}{c}{Upscale Ratio}\\

			  & $\times$1.5 & $\times$2 & $\times$3.5 & $\times$4   & $\times$5.5 & $\times$6 & $\times$7.5 & $\times$8  \\ 
			\midrule
   			  MISF\ding{213}BI  &29.53/0.970 & 27.46/0.935 &26.11/0.905 & 25.55/0.871  &25.56/0.811 &24.65/0.799 &24.34/0.805& 23.94/0.758\\

    		MISF\ding{213}EDSR	&\---   & 27.30/0.921 &\--- &  26.83/0.872 &\---& 25.39/0.817&\---& 25.43/0.799\\
  
      		MISF\ding{213}LTE	&27.41/0.892&    26.40/0.848 &26.11/0.805 & 26.09/0.829 &25.71/0.768 &25.38/0.766 &24.58/0.760 & 24.35/0.758\\
 
        		MISF\ding{213}ELAN	&\---    & 26.71/0.909 &\--- & 26.89/0.873 &\--- &25.77/0.825 &\---& 24.79/0.785 \\
  

   			  BI\ding{213}MISF &27.04/0.949 & 27.02/0.937& 26.42/0.892 & 26.28/0.878 &25.90/0.844 &25.64/0.832 &24.83/0.805&24.32/0.794\\

    		EDSR\ding{213}MISF	&\---   & 24.62/0.910 &\--- &  22.83/0.840&\---& 22.97/0.805&\---& 22.47/0.773\\
  
      		LTE\ding{213}MISF	&22.92/0.866&    22.90/0.849 &22.85/0.800 & 22.65/0.781 &21.78/0.725 &21.65/0.714 &21.43/0.692 & 21.34/0.686\\
 
        		ELAN\ding{213}MISF	&\---   & 24.23/0.897 &\--- &  21.06/0.766&\---& 20.54/0.717&\---& 19.93/0.680\\

			  LTE &26.05/0.865 & 29.08/0.906  &28.19/0.854 &28.82/0.872 &28.21/0.824 &28.37/0.817& 27.76/0.804 & 27.70/0.799\\

     \midrule
			DEAR  &\textbf{33.11}/\textbf{0.986}& \textbf{31.32}/\textbf{0.977} &\textbf{30.07}/\textbf{0.971}& \textbf{29.84}/\textbf{0.968} &\textbf{29.13}/\textbf{0.932}& \textbf{29.13}/\textbf{0.924}&\textbf{28.93}/\textbf{0.887}& \textbf{28.19}/\textbf{0.859}\\

		\bottomrule
	\end{tabular}}

\vspace{5pt}

\caption{Results from the DEAR and baselines on CelebAHQ~\cite{karras2017progressive} dataset under different upscale ratios. We use PSNR/SSIM to evaluate the performance.}
\label{result-SR}

\end{table*}

%
\begin{figure*}[t]
  \includegraphics[width=1\textwidth]{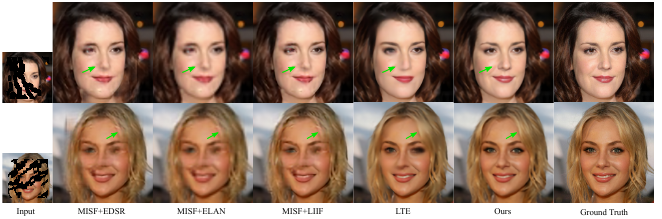} 
  \vspace{-5pt}
  \caption{Qualitative results on CelebAHQ~\cite{karras2017progressive} dataset.}
  \label{img-celebA}
\end{figure*}
%

{\bf Datasets.}
Since \textit{SuperInpaint} is the new proposed task, no previous dataset can be used directly in our setting.
We modify three widely used datasets CelebAHQ~\cite{karras2017progressive}, Places365~\cite{zhou2017places}, and FFHQ~\cite{karras2019style}. 
CelebAHQ is a large-scale face image dataset that contains 30,000 high-resolution face images selected from the CelebA dataset~\cite{liu2015faceattributes}. 
Places365 is a huge scene dataset that contains 1.8 million images from 365 scene categories.
FFHQ is composed of 70,000 high-quality human images which are variable in terms of age, ethnicity, and background.
To get the corrupted LR image, we down-sample the HR image and add the mask from the irregular mask dataset~\cite{liu2018image}.

{\bf Baselines.}
We combine the inpainting models (MISF~\cite{li2022misf} and LAMA~\cite{suvorov2021resolution}) and SR models (Bilinear Interpolation(BI), EDSR~\cite{lim2017enhanced}, LTE~\cite{lte-jaewon-lee}, and ELAN~\cite{zhang2022efficient}) to generate baselines as discussed in Sec.~\ref{subsec:limitations}.
The inpainting models are trained for LR and HR images separately to fit different input sizes.
We also test the performance of LTE on \textit{SuperInpaint} task directly.

{\bf Evaluation metrics.}
We choose four widely used image quality metrics PSNR (peak signal-to-noise ratio), SSIM (structural similarity index), L1, and LPIPS~\cite{zhang2018perceptual} (perceptual similarity) to evaluate the performance of all networks. 
PSNR, SSIM, and L1 reflect the quality of the generated image. 
LPIPS measures the perceptual distance between the restored image and ground truth.

\subsection{Quantitative Results}

We show the results from DEAR and baselines on CelebAHQ, Places365, and FFHQ datasets in Table~\ref{main-result}. 
Proposed method earns the best PSNR, SSIM, L1, and LPIPS scores compared with all baselines. 
On CelebAHQ dataset, compared with MISF\ding{213}LTE, our model DEAR gets 3.75 higher PSNR and improves the SSIM by 13.9\%.
Compared with LTE, DEAR can improve the PSNR by 1.58 and increase the SSIM by 11.3\%.
On Places365 dataset, compared with LAMA\ding{213}LTE, DEAR boosts the PSNR by 3.68.
DEAR brings 10.2\% SSIM improvement in contrast to LTE.
On FFHQ dataset, our model lowers the L1 by 0.003 and LPIPS by 0.086 in contrast to LTE. 
Those results show that the model combination cannot address the \textit{SuperInpaint} challenge. 
DEAR combines the merits from inpainting and SR tasks and earns promising results, since it preserves more high-frequency information and reconstructs the corrupted area with the help of the unmasked area information and importance map.

In Table~\ref{result-SR}, we show the reconstruction ability of DEAR and baselines under different upscale ratios.
DEAR and LTE~\cite{lte-jaewon-lee} are trained on $\times$4 setting.
ELAN~\cite{zhang2022efficient} and EDSR~\cite{lim2017enhanced} need to be trained separately on each setting.
On $\times$2 setting, DEAR improves PSNR by 4.02 and SSIM by 5.6\% when compared with MISF\ding{213}EDSR.
On $\times$8 setting, our model brings 0.49 PSNR improvement in contrast to LTE. 
We can see that DEAR earns the best performance under all ratios with the help of implicit neural function and effective model components, while proposed baselines suffer from poor performance and upscale ratio restriction.

\subsection{Qualitative Results}

\begin{figure*}[ht]
  \includegraphics[width=\textwidth]{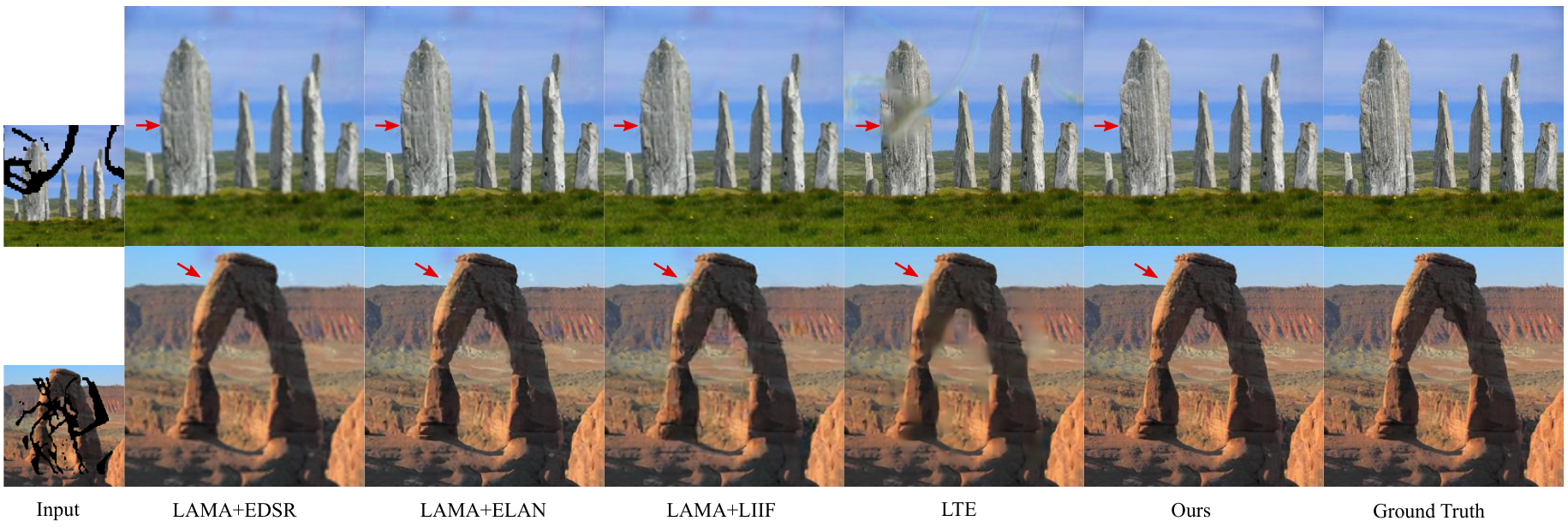} 
  \vspace{-5pt}
  \caption{Qualitative results on Places365~\cite{zhou2017places} dataset.}
  \label{img-p2}
\end{figure*}

To further justify the ability of DEAR, we show qualitative results from our model and baselines.
The qualitative results on CelebAHQ dataset are shown in Figure~\ref{img-celebA}, and the results from Places365 dataset are shown in Figure~\ref{img-p2}. 
Those baselines are unable to generate reliable results because SR models will amplify the flaws in reconstructed LR image. 
LTE cannot restore the corrupted area correctly, since it doesn't learn enough information from unmasked regions.
Compared with those methods, our method generates more high-fidelity outputs.
As illustrated in the second row of Figure~\ref{img-celebA}, DEAR preserves more texture information in the hair area. 
As shown in the first row of Figure~\ref{img-p2}, DEAR generates more details in the stone part. 

%
%

\subsection{Ablation Study}

\paragraph{SR capability of DEAR.}
DEAR can be used on the traditional SR task directly, where the input is the clean low-resolution image.
To justify the SR ability of DEAR, we compare its results with LIIF~\cite{chen2021learning} and LTE~\cite{lte-jaewon-lee} under several upscale ratios.
All models are trained under $\times$4 setting.
As shown in Table~\ref{ablation-SR}, 
compared with LIIF, DEAR boosts the PSNR by 0.48 on $\times$2 setting and highlights SSIM to 82.6\% on $\times$8 setting.
 DEAR can optimize the PSNR by 1.9 on $\times$2 setting in contrast to LTE.
Proposed network uses high-pass filters to preserve more high-frequency information which plays a key role in image reconstruction.

\begin{table}[t]
	\centering
	\renewcommand\arraystretch{1.0}
	\footnotesize
	\setlength{\tabcolsep}{0.8mm}{
		\begin{tabular}{l|ccccc}
			\toprule
   
                 \multirow{2}{*}{Method} &  \multicolumn{4}{c}{Upscale Ratio}\\

			  & $\times$2 & $\times$4 & $\times$6 & $\times$8 \\ 
	\midrule
			  LIIF & 34.32/0.963  & 30.98/0.882 &30.13/0.838 & 29.83/0.819\\
            LTE & 32.90/0.955  & \textbf{31.77}/ \textbf{0.897} & \textbf{30.93}/0.841 & 30.15/0.821\\
			DEAR  & \textbf{34.80}/\textbf{0.976} & 31.39/0.890 & 30.48/\textbf{0.846}& \textbf{30.17}/\textbf{0.826}\\

		\bottomrule
	\end{tabular}}

\vspace{0pt}

 \caption{Study on SR ability of DEAR on FFHQ~\cite{karras2019style} dataset. We use PSNR/SSIM to evaluate the performance.}
\label{ablation-SR}
\end{table}

\paragraph{Inpainting capability of DEAR.}
DEAR can also be used on inpainting task directly, where the input and output have the same resolution.
The results are shown in Table~\ref{ablation-inpaint}.
On the CelebAHQ dataset, compared with MISF, DEAR raises the PSNR by 0.21. On the Places365 dataset, our model can increase the PSNR by 0.59 in contrast to MISF.
Those results show that DEAR also has good inpainting ability even compared with SOTA inpainting models.
The unmask-attentional semantic embedding and pixel-wise importance map guarantee that our method has a high generalization capability.

\begin{table}[t]
	\centering

	\renewcommand\arraystretch{1.0}
	\footnotesize
	\setlength{\tabcolsep}{0.6mm}{
		\begin{tabular}{l|cccc|cccc}
			\toprule
    \multirow{2}{*}{Method} &  \multicolumn{4}{c|}{CelebAHQ} &  \multicolumn{4}{c}{Places365}\\

               & PSNR$\uparrow$  & SSIM$\uparrow$  &   L1 $\downarrow$ &  LPIPS$\downarrow$ & PSNR$\uparrow$  & SSIM$\uparrow$  &   L1 $\downarrow$ &  LPIPS$\downarrow$ \\ 
	\midrule
			  LAMA & 33.36  & 0.990 &0.011 & 0.012&28.26& 0.962 & 0.025& 0.047\\
            MISF & 34.28  & \textbf{0.991} &0.012 &\textbf{0.011}&30.09& \textbf{0.969} & 0.022&0.041\\

			DEAR  & \textbf{34.49} & 0.974 & \textbf{0.010}& 0.017& \textbf{30.68} &0.950 & \textbf{0.021} & \textbf{0.039}\\

		\bottomrule
	\end{tabular}}

\vspace{0pt}

 \caption{Study on inpainting ability of DEAR.}
\label{ablation-inpaint}
\end{table}

\paragraph{Model components.}
To show that all model components aid the performance improvement, we do ablation experiments by adding each DEAR part iteratively.
As shown in Table~\ref{ablation}, on CelebAHQ dataset, adding detailed-enhanced semantic embedding (DSE) can improve the PSNR by 0.95 and improve SSIM by 2.2\%.
Those results show the necessity of high-frequency information.
Adding unmask-attentional semantic embedding (USE) can bring 0.77 PSNR improvement and lower the L1 by 0.007. The pixel-wise importance map (PIM) can help the model highlight the SSIM by 1.5\% meanwhile lowering LPIPS by 0.034.

\begin{figure}[t]
  \includegraphics[width=0.47\textwidth]{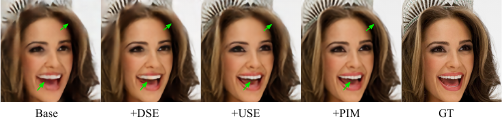} 
  \vspace{0pt}
  \caption{Outputs from DEAR variants after adding detailed-enhanced semantic embedding  (DSE), unmask-attentional semantic embedding  (USE), and pixel-wise importance map (PIM).}
  \label{abla-img}
\end{figure}

We also show the output from each model variant in Figure~\ref{abla-img}. We can see that the base model gets a poor result, leaving many blurred parts. The high frequency feature helps the model recover more detail information. 
The models with USE and PIM can reconstruct more realistic images.

\begin{table}[t]
	\centering
	
	\renewcommand\arraystretch{1.0}
	\footnotesize
	\setlength{\tabcolsep}{0.6mm}{
		\begin{tabular}{l|cccc|cccc}
			\toprule
			   \multirow{2}{*}{Variant} &  \multicolumn{4}{c|}{CelebAHQ} &  \multicolumn{4}{c}{Places365}\\
             &  PSNR$\uparrow$  & SSIM$\uparrow$  &   L1 $\downarrow$ &  LPIPS$\downarrow$&  PSNR$\uparrow$  & SSIM$\uparrow$  &   L1 $\downarrow$ &  LPIPS$\downarrow$  \\
			\midrule
		Base  & 27.21 & 0.916 & 
               0.035 & 0.121 & 23.93& 0.755&0.046& 0.239 \\
               +DSE   & 28.16& 0.948 &
               0.030 & 0.094  &  24.66& 0.786&0.042&0.224\\
               +USE  & 28.93 & 0.953 & 
               0.023 & 0.087  & 25.01& 0.793 &0.043& 0.213\\
               +PIM   & \textbf{29.84} & \textbf{0.968} & 
               \textbf{0.022} & \textbf{0.053}& \textbf{25.61} &\textbf{0.824}& \textbf{0.040}&\textbf{0.170}  \\

			\bottomrule
	\end{tabular}}
 \vspace{0pt}
 \caption{Ablation study on DEAR components by adding detailed-enhanced semantic embedding (DSE), unmask-attentional semantic embedding (USE), and pixel-wise importance map (PIM).}
 \label{ablation}
\end{table}

\subsection{Difference to Existing Implicit Representations}

\begin{figure}[t]
  \includegraphics[width=0.47\textwidth]{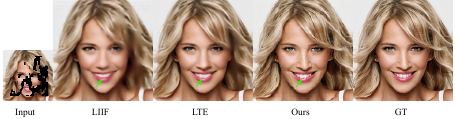} 
  \vspace{0pt}
  \caption{Outputs from LIIF~\cite{chen2021learning}, LTE~\cite{lte-jaewon-lee}, and DEAR  on \textit{SuperInpaint} task.}
  \label{analysis-img}
\end{figure}

In this part, we analyze the performance of LIIF~\cite{chen2021learning} , LTE~\cite{lte-jaewon-lee} and DEAR on \textit{SuperInpaint} task.
All three models can be used to restore the images in arbitrary higher resolutions.
The quantitative results are shown in Table~\ref{analysis}, and qualitative results are shown in Figure~\ref{analysis-img}.
LIIF uses implicit neural representation to reconstruct arbitrary resolution images firstly, but it ignores some high-frequency information due to MLP characteristics.
To overcome this drawback, LTE adds high-frequency embeddings in Fourier domain. 
 DEAR uses a high-pass filter to enhance the high-frequency feature.
As shown in Table~\ref{result-SR} and Table~\ref{ablation-SR}, LTE faces significant performance drop in the low-upscale ratios, while DEAR maintains high performance under all ratios.
LIIF and LTE are unable to learn sufficient information from the unmasked region, leaving many blurred areas.
We further propose unmask-attentional semantic embedding to suppress the undesired pixel features. We generate pixel-wise importance map to tell the model whether certain pixels should be considered or not.
Those components help DEAR generate high-fidelity image in arbitrary resolutions.

\begin{table}[t]
	\centering
 
	\renewcommand\arraystretch{1.0}
	\footnotesize
	\setlength{\tabcolsep}{0.6mm}{
		\begin{tabular}{l|cccc|cccc}
			\toprule
    \multirow{2}{*}{Method} &  \multicolumn{4}{c|}{CelebAHQ} &  \multicolumn{4}{c}{Places365}\\

               & PSNR$\uparrow$  & SSIM$\uparrow$  &   L1 $\downarrow$ &  LPIPS$\downarrow$ & PSNR$\uparrow$  & SSIM$\uparrow$  &   L1 $\downarrow$ &  LPIPS$\downarrow$ \\ 
	\midrule
			  LIIF & 27.91  & 0.841 &0.029 & 0.252&23.38& 0.716 & 0.047& 0.356\\
     
            LTE & 28.26  & 0.855 &0.025 &0.236&23.87& 0.722 & 0.041&0.354\\

			DEAR  & \textbf{29.84} & \textbf{0.968} & \textbf{0.022}& \textbf{0.053}& \textbf{25.61} &\textbf{0.824} & \textbf{0.040} & \textbf{0.170}\\
		\bottomrule
	\end{tabular}}
\vspace{0pt}
 \caption{Comparisons among LIIF~\cite{chen2021learning}, LTE~\cite{lte-jaewon-lee}, and DEAR on \textit{SuperInpaint} task.}
\label{analysis}
\end{table}

\section{Conclusions}

In this paper, we identified a challenging image restoration task named \emph{SuperInpaint}. Given a low-resolution image with large missing regions, \emph{SuperInpaint} is to fill in the missing pixels while generating high-fidelity images with arbitrarily higher resolutions.
We proposed the detail-enhanced attentional implicit representation (DEAR) to achieve this task with three new designs including detail-enhanced semantic embedding extraction, unmask-attentional semantic embedding extraction, and pixel-wise importance map extraction.
We also built 18 baseline methods based on state-of-the-art inpainting and super-resolution methods and constructed three new datasets for \emph{SuperInpaint} based on existing public datasets. 
The extensive results demonstrate that DEAR outperforms all baselines significantly.

{\small
\bibliographystyle{ieee_fullname}
\bibliography{egbib}
}

\end{document}